\pdfoutput=1

\documentclass[11pt]{article}

\usepackage[final]{acl}

\usepackage{times}
\usepackage{latexsym}

\usepackage[T1]{fontenc}

\usepackage[utf8]{inputenc}

\usepackage{microtype}

\usepackage{inconsolata}

\usepackage{graphicx}
\usepackage{rotating}
\usepackage{adjustbox}
\usepackage{amsmath}
\usepackage{multirow}
\usepackage{arydshln}

\title{QPaug: Question and Passage Augmentation for Open-Domain Question Answering of LLMs}

\author{
 \textbf{Minsang Kim\textsuperscript{1}},
 \textbf{Cheoneum Park\textsuperscript{2}},
 \textbf{Seungjun Baek\textsuperscript{1}}\thanks{Corresonding Author}
\\
 \textsuperscript{1}Korea University,
 \textsuperscript{2}Hanbat National University,
\\
 \texttt{kmswin1@korea.ac.kr}
}

\begin{document}
\maketitle
\begin{abstract}
Retrieval-augmented generation (RAG) has received much attention for Open-domain question-answering (ODQA) tasks as a means to compensate for the parametric knowledge of large language models (LLMs).
While previous approaches focused on processing retrieved passages to remove irrelevant context, they still rely heavily on the quality of retrieved passages which can degrade if the question is ambiguous or complex. In this paper, we propose a simple yet efficient method called question and passage augmentation (QPaug) via LLMs for open-domain QA. QPaug first decomposes the original questions into multiple-step sub-questions. By augmenting the original question with detailed sub-questions and planning, we are able to make the query more specific on what needs to be retrieved, improving the retrieval performance. In addition, to compensate for the case where the retrieved passages contain distracting information or divided opinions, we augment the retrieved passages with self-generated passages by LLMs to guide the answer extraction. Experimental results show that QPaug outperforms the previous state-of-the-art and achieves significant performance gain over existing RAG methods. The source code is available at \url{https://github.com/kmswin1/QPaug}.
\end{abstract}

\section{Introduction}
Large language models (LLMs) have shown remarkable in-context learning capability~\cite{brown2020language, touvron2023llama} for various real-world applications such as assistant chatbot~\cite{achiam2023gpt, team2023gemini}, robot planning~\cite{wang2023gensim}, search ranking~\cite{sun2023chatgpt}, and code generation~\cite{chen2021evaluating}. However, the knowledge of LLMs is limited to the pre-training corpus, making it difficult to provide answers to questions on up-to-date information. 
To overcome such limitations via external knowledge sources, recent works have focused on the retrieval-augmented-generation~(RAG) ~\cite{guu2020retrieval, lewis2020retrieval}. The RAG systems consist of \emph{retrievers} which search and retrieve related information from knowledge sources and \emph{readers} which generate responses based on the retrieved information.
The RAG approach has been proven effective in leveraging external knowledge to complement the parametric knowledge of LLMs.

Open-domain Question Answering~(ODQA) \cite{yang2018hotpotqa, kwiatkowski2019natural, ho2020constructing, karpukhin2020dense} is one of the NLP tasks most relevant to RAG systems. In the ODQA, retrievers search for relevant passages from questions, and readers answer the questions based on retrieved contexts. Research efforts have been put into enhancing retrievers~\cite{xiong2020approximate, izacard2021unsupervised} and readers~\cite{kenton2019bert, lewis2020bart}. For example, RAPTOR \cite{sarthi2023raptor} is an advanced retriever which recursively captures multiple levels of details of a text using a tree. However, even advanced retrievers may fetch poor passages when \emph{the question is ambiguous or complex}, requiring question clarification~\cite{zamani2020generating, lee2023asking, kim2023tree} or multi-step retrieval~\cite{feldman2019multi, welbl2018constructing, trivedi2022musique}, rendering the retrieval inaccurate and inefficient.

Thus, for tough questions, it is unclear whether the retrieved passages are of high quality, i.e., they contain context relevant to the question, which makes the design of readers challenging. The problem can be alleviated if the parametric knowledge of LLMs can be tuned to extract answers given the relevant retrieved passages~\cite{guu2020retrieval, izacard2021leveraging}. However, fine-tuning is often infeasible due to the sheer scale of LLMs, or some LLMs are essentially black-box APIs~\cite{team2023gemini, achiam2023gpt}. Recently, \cite{lazaridou2022internet, kim2023sure} have proposed to process retrieved passages via {LLM prompting} without fine-tuning, but with Internet retrieval and reranking~\cite{lazaridou2022internet}, or with summarization and verification~\cite{kim2023sure}. However, these approaches are still limited in that they heavily rely on the quality of contexts provided by retrieved passages.

In this paper, we deal with question complexity and complement the quality of retrieval by in-context learning. The goal is to harmoniously combine the parametric and non-parametric knowledge of LLMs through prompting. We propose QPaug (pronounced \emph{cue-pug}) which stands for question and passage augmentation. Firstly, the question augmentation is based on the hypothesis that the LLMs can decompose complex questions into multiple easier sub-questions~\cite{kojima2022large}. The sub-questions contain fine-grained information and planning as to what knowledge should be retrieved from external sources. The question for retrieval is composed by augmenting the original question with the sub-questions. Secondly, the passage augmentation is based on the hypothesis that the LLMs' parametric knowledge is the most pragmatic alternative source to non-parametric (retrieved) knowledge to deal with the degraded quality of retrieval. A self-generated passage is composed with respect to the augmented question, leveraging extensive knowledge of LLMs. Also, self-generation can prevent irrelevant retrieved passages from overriding the LLMs' factual knowledge~\cite{zheng2023can, wei2023larger}. We augment the retrieved passages with the self-generated passages. Experiments show that the question and passage augmentation through LLMs complement the retrieved generation well, achieving state-of-the-art performance on ODQA benchmark datasets. 

Our contributions are summarized as follows. (i) We propose a simple yet effective framework for the LLM-based question and passage augmentation method~(QPaug) based on the prompting of LLMs. (ii) QPaug successfully decomposes and augments questions improving the performance of retrieval. In addition, the self-generated passage by QPaug, when combined with retrieved passages, strengthens the factual knowledge for the reader, achieving large performance gains. (iii) Extensive ablation studies show that QPaug can be integrated with various LLMs and retrievers, exhibiting excellent performances on several ODQA benchmarks.

\begin{figure*}[t!]
    \centering
    \includegraphics[width=.9\textwidth]{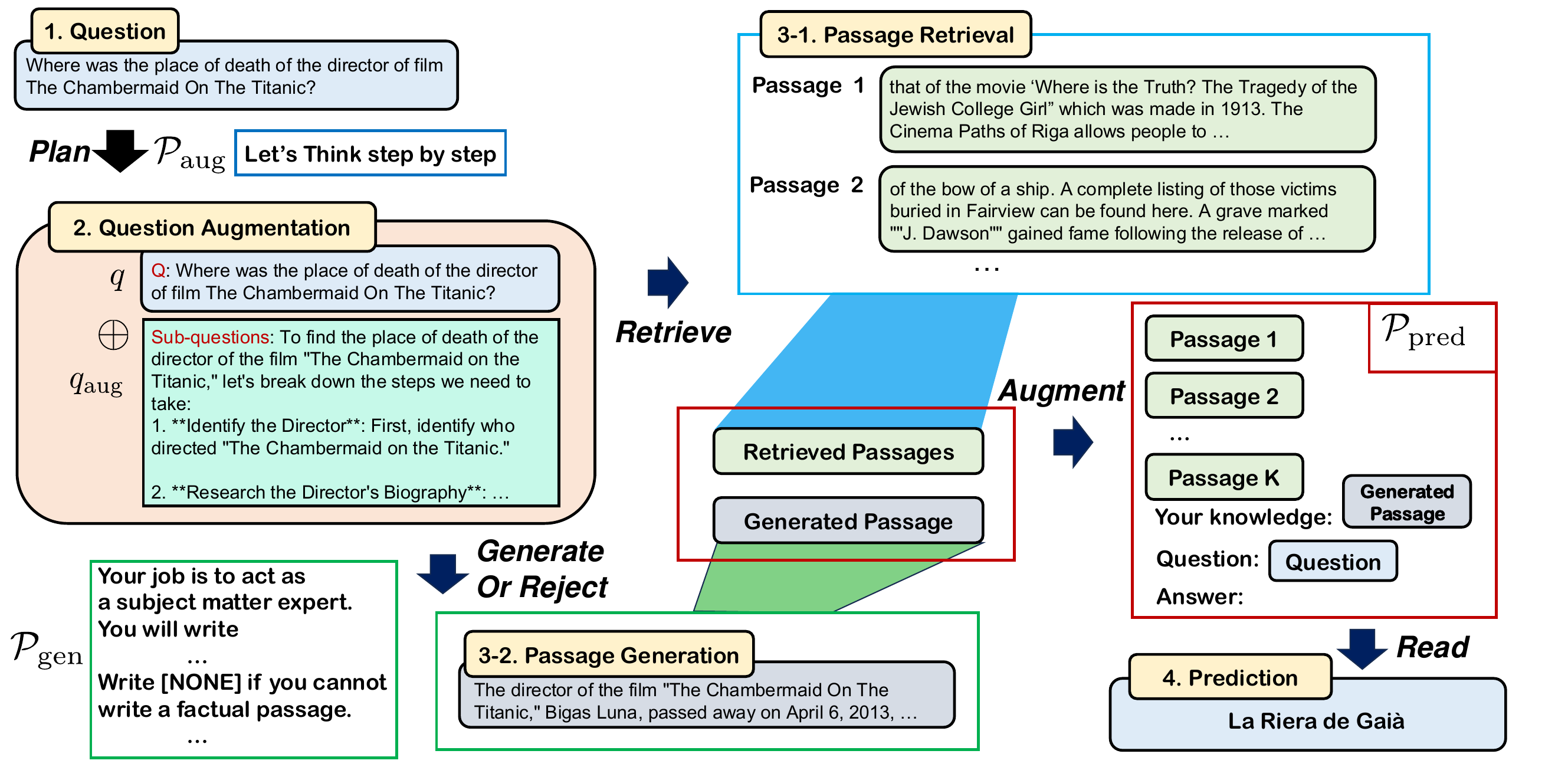}
    \caption{\textbf{Step 1}. LLMs receive questions. \textbf{Step 2}. The original question is decomposed into multi-step sub-questions on what information should be retrieved. \textbf{Step 3-1}. Retrieve passages from augmented questions. \textbf{Step 3-2}. Self-generate as a passage from the augmented questions using factual knowledge. The LLM is asked to generate \texttt{[NONE]} if it does not have the required knowledge. \textbf{Step 4.} Augment retrieved passages with the self-generated passage, then predict answers. $\mathcal{P}_x$ denotes the prompt at each step.}
    \label{fig:model}
\end{figure*}
\section{Related Work}
\subsection{Open-Domain Question Answering} \cite{chen2017reading} first proposed retrieve-and-read system for solving open-domain question answering tasks. 
Following conventional lexical-based sparse retriever systems like BM25~\cite{robertson2009probabilistic},  DPR~\cite{karpukhin2020dense} proposed a \emph{dense passage retrieval} for a semantic retriever system. The semantic retriever is based on sentence embeddings, and there have been a number of works for improving embeddings~\cite{reimers2019sentence, xiong2020approximate, izacard2021unsupervised, wang2022text, gao2023precise}. By contrast, {the reader system which extracts answers from retrieved documents} consists of extractive methods such as BERT~\cite{kenton2019bert} or RoBERTa~\cite{liu2019roberta} and generative methods like BART~\cite{lewis2020bart} or T5~\cite{raffel2020exploring}.

\subsection{Retrieval-Augmented Generation}
Augmenting language models with retrieved information from external knowledge sources have proven effective for a wide range of NLP tasks~\cite{guu2020retrieval, lewis2020retrieval, borgeaud2022improving}. In the LLM era, \cite{lazaridou2022internet, izacard2023atlas} proposed an in-context learning-based retrieval-augmented generation methods. \cite{asai2023self} proposed Self-RAG which generates and reflects on retrieved passages and own generated text using reflection tokens. \cite{sarthi2023raptor} proposed recursive retrieval methods using embedding, clustering, and summarizing chunks of text, where they construct a tree with differing levels of summarization. \cite{kim2023sure} proposed summarizing retrieved passages conditioned on candidate answers to select the more relevant context. 

\subsection{Prompting of Large Language Models}
GPT-3~\cite{brown2020language} opened a few-shot learning era of language models. \cite{si2022prompting} extensively studied about prompting of GPT-3 using manually designed prompts on diverse tasks. They showed that GPT-3 is more reliable with proper prompts. \cite{wei2022chain, kojima2022large} proposed chain-of-thoughts~(CoT), which decomposes a problem into multi-step subproblems.  
In addition, there have been modified works of CoT~\cite{wang2023chain, dhuliawala2023chain, yao2024tree, besta2024graph}. \cite{wang2023towards} extensively studied the properties of CoT, and illustrated that even incorrect reasoning paths can improve performances of LLM reasonings.

\section{Method}
\subsection{Problem Statement and Notations}
Open domain question answering (ODQA) is the extension of QA tasks in which explicit evidence or context is not provided to the model. Thus, ODQA requires other knowledge sources such as an external knowledge base. The basic strategy to solve ODQA tasks is \emph{retrieve-and-read}. Specifically,  \emph{retriever} $\mathcal{R}$ obtains top-$K$ related passages $\mathcal{C}_K$ from knowledge source $\mathcal{Z}$:
\begin{align}
    &\mathcal{C}_K = \mathcal{R}(q, \mathcal{Z}, K)
\end{align}
Then \emph{reader} extracts predicted answer $\hat{a}$ about question $q$ using given the retrieved passages $\mathcal{C}_K$:
\begin{align}
    \hat{a} = \mathcal{M}(\mathcal P(q, \mathcal{C}_K))
\end{align}
where we denote the LLM-generated text as $\mathcal{M}$($\cdot$), and $\mathcal P$ denotes certain prompt.

\subsection{Question Augmentation}
We propose \emph{question augmentation} to enhance the retrieval capability. 
The original question is decomposed into multiple sub-questions which can arise during the reasoning process. The motivation is that a better retrieval is possible if we augment the original question with those sub-questions which can provide fine-grained hints for resolving the question. We utilize a zero-shot chain of thoughts (CoT)~\cite{kojima2022large} for composing sub-questions. Specifically, we use prompt $\mathcal P_\textrm{aug}(\cdot)$ for question $q$, where $\mathcal P_\textrm{aug}(q)$ adds the sentence \texttt{``Let's think step-by-step''} \cite{kojima2022large} to $q$. $\mathcal P_\textrm{aug}(q)$ is then passed to LLM to generate augmented question $q_\textrm{aug}$, i.e.,
\begin{equation}   
  q_\textrm{aug} = \mathcal{M}(\mathcal P_\textrm{aug}(q))\\
\end{equation}
An example is given in Fig.\ \ref{fig:model}: see Steps 1 and 2. In Step 2, we observe that $q_\textrm{aug}$ contains fine-grained instructions to tackle the question. Notably, although prior zero-shot CoT \cite{kojima2022large} is shown to achieve large performance gains on arithmetic reasoning tasks, its effectiveness was less examined on ODQA tasks. However, we observe that augmenting the question with reasoning steps planned out by CoT boosts the retrieval capability, leading to improved performances on ODQA tasks.

Next, the augmented question is used for retrieval. Specifically, we create query $\hat q$ by concatenating $q$ and $q_\textrm{aug}$:
\begin{equation}
    \hat{q} = q \oplus q_\textrm{aug}
\end{equation}
where $\hat{q}$ is used for the retrieval. We retrieve top-$K$ passages $c_1,...,c_K$ using Approximated Max Inner Product Search~(MIPS)~\cite{ram2012maximum, auvolat2015clustering}.
\begin{equation}
    \mathcal{C}_{K}  = \mathcal{R}(\hat{q},\mathcal{Z}, K) = \{c_1,...,c_K\}.
\end{equation}

\subsection{Passage Self-Generation}
We propose to utilize the vast knowledge of LLMs to complement the contextual information provided by retrieved passages. Specifically, the LLM is asked to self-generate a passage regarding the question. We augment the retrieved passages with the generated passage. 
Let $\mathcal P_\textrm{gen}(q)$ denote a prompt for passage generation given question $q$. We have
\begin{align}
    \hat{c} = \mathcal{M}(\mathcal P_\textrm{gen}(\hat{q}))
\end{align}
where $\hat{c}$ denotes the LLM-generated passage. For example, see  Step 1 and 2-2 in Fig.\ \ref{fig:model}, and see the outline of prompt $\mathcal P_\textrm{gen}$ for passage generation in Step 2-2. 

Finally, we extract predicted answers from the LLM using $\mathcal{C}_K$ and $\hat{c}$. The final prompt $\mathcal P_\textrm{pred}$ for prediction contains $q$, $\mathcal{C}$, and $\hat{c}$, which is input to LLM for the predicted answer, i.e.,
\begin{equation}
    \hat{a} = \mathcal{M}(\mathcal P_\textrm{pred}(q, \mathcal{C}_K, \hat{c}))
\end{equation}

An important consideration is that, the LLM may not have sufficient knowledge for the question, and should refrain from generating plausible but incorrect passages. To derive as truthful passages as possible, we explicitly instruct LLM to base its passage on factual knowledge, and to generate \texttt{[NONE]} in case it does not have sufficient knowledge: see the detailed prompt $\mathcal P_\textrm{gen}$ for passage generation in Table~\ref{fig:prompt_palm} of Appendix~\ref{appendix:prompt_templates}. There are two cases as a result.

\noindent\textbf{LLM admits lack of knowledge.} In this case, LLM returns \texttt{[NONE]} passage, and only the retrieved passages are used at the final \emph{read} step to predict the answer. 

\noindent\textbf{LLM hallucinates.} LLM provides a bogus passage, believing that it is from factual knowledge. To prevent possible hallucinations from mixing with retrieved passages, we label the LLM-generated passage as \texttt{``Your Knowledge:''} in composing the final prompt $\mathcal P_\textrm{pred}$ for the answer prediction: see the box of $\mathcal P_\textrm{pred}$ above Step 4 in Fig.\ \ref{fig:model}. Presumably, if a majority of $K$ retrieved passages provide relevant information, LLM is likely to recover the correct answer in spite of hallucination. Examples of the above two cases are provided in Table~\ref{tab:pgen_none} and Table~\ref{tab:pgen_wrong} of Appendix \ref{appendix:PALM}. Overall, the proposed passage augmentation is more beneficial than harmful, as demonstrated by experiments in the following section. 

\begin{table}[t!]
    \centering
    \begin{tabular}{|l|l|}
\hline
\textbf{Dataset} & \textbf{\# samples} \\ \hline
Passages         & 21,015,324                    \\
\hline
NQ               & 4,289                \\ 
2wiki            & 12,576               \\ 
Hotpot           & 7,405                \\ \hline
\end{tabular}
    \caption{Dataset Statistics of passages and test sets of ODQA benchmark datasets.}
    \label{tab:dataset_statistics}
\end{table}

\section{Experiment}

\subsection{Experimental settings}
\textbf{Evaluation datasets and metrics.} We experiment with zero-shot QA tasks on three ODQA benchmarks: Natural Questions~(NQ)~\cite{kwiatkowski2019natural}, 2wiki hop questions~(2wiki)~\cite{ho2020constructing}, and HotpotQA~(Hotpot)~\cite{yang2018hotpotqa}. The test sets from those datasets are used for the experiments.2wiki and HotpotQA are multi-hop QA datasets that typically require multi-step retrieval for traditional methods.
As the knowledge sources for retrieval, we use 21M passages of Wikipedia dump proposed by DPR~\cite{karpukhin2020dense}. Table~\ref{tab:dataset_statistics} shows the dataset statistics of passages and test sets.
Since the datasets include both short and long answers in benchmark datasets, answers are not only words but also phrases or sentences. Thus, we use Rouge-L and F1 scores as evaluation metrics to compute fine-grained scores the same as~\cite{nguyen2016ms, kwiatkowski2019natural, ho2020constructing, yang2018hotpotqa}. We normalize answers and predictions before computing both metrics similar to~\cite{rajpurkar2016squad}. 
All the baseline methods use $K=10$ retrieved passages as the input to the reader. For a fair comparison, QPaug uses $K=9$ retrieved passages combined with one LLM-generated passage. 
\noindent \textbf{Baseline models.} We experiment with three  retrievers: SBERT~\cite{reimers2019sentence}, ANCE~\cite{xiong2020approximate}, and Contriever~\cite{izacard2021unsupervised}. In addition, we experiment with three LLMs as readers: Llama-2~\cite{touvron2023llama}, GPT-3.5~\cite{ouyang2022training} and GPT-4~\cite{achiam2023gpt}. We use the same LLM as the reader model for the proposed question and passage augmentation. We compare QPaug with no retrieval methods, e.g., chain-of-thoughts~\cite{kojima2022large} and Self-verification~\cite{weng2023large} as well as context-augmentation methods such as Rerank~\cite{lazaridou2022internet} and SuRE~\cite{kim2023sure}.

\begin{table}[h!]
    \centering
    \begin{tabular}{|c|c|c|c|}
        \hline
        \textbf{Method} & \textbf{NQ}  & \textbf{2wiki} & \textbf{Hotpot}  \\ \hline
        No retrieval  & 37.9 & 27.1  & 35.3  \\
        Chain-of-thoughts  & 38.2 & 28.2 & 35.5  \\
        Rerank  & 38.0 & 26.6 & 33.2 \\
        Self-verification & 38.4 & 30.8 & 35.9 \\
        SuRE & 40.4 & 32.6 & 33.6  \\
        \hline
        \textbf{QPaug (Proposed)} & \textbf{44.6} & \textbf{35.5} & \textbf{45.1}  \\ \hline
    \end{tabular}
    \caption{Comparison between QPaug and baseline methods. We use retriever as Contriever for Rerank, SuRE, and QPaug, and use GPT-3.5 as the LLM of all methods. The evaluation metric is the F1 score between answers and predictions. \textbf{Bold} indicates the best performance.}
    \label{tab:comparison_sure}
\end{table}

\begin{table*}[t!]
\begin{adjustbox}{width=1.\textwidth, center}
\begin{tabular}{|l|l|l|ll|ll|ll|}
\hline
LLM    & Datasets & No retrieval & SBERT       & \textbf{+ QPaug}   & ANCE        & \textbf{+ QPaug}    & Contriever  & \textbf{+ QPaug}\\ \hline
\multirow{4}{*}{GPT-4}   & NQ       & 37.0 / 42.0  & 42.8 / 46.0 & \textbf{43.7 / 54.0} & 40.8 / 45.0 & \textbf{43.9 / 52.0} & 39.4 / 43.7 & \textbf{42.4 / 52.0} \\
                         & 2wiki    & 26.8 / 37.6  & 27.8 / 36.5 & \textbf{37.3 / 49.1} & 25.0 / 33.6 & \textbf{36.7 / 47.2} & 24.6 / 32.6 & \textbf{35.2 / 47.2} \\
                         & Hotpot   & 33.9 / 42.2  & 39.2 / 49.0 & \textbf{44.9 / 54.6} & 34.6 / 43.6 & \textbf{42.4 / 52.0} & 34.0 / 43.2 & \textbf{43.7 / 53.5} \\ \cline{2-9} 
                         & Average  & 32.6 / 40.6  & 36.6 / 43.8 & \textbf{42.0 / 52.6}   & 33.5 / 40.7 & \textbf{41.0 / 50.4}       & 32.7 / 39.8 & \textbf{40.4 / 50.9}                           \\ \hline
\multirow{5}{*}{GPT-3.5} & Datasets & No retrieval & SBERT       & \textbf{+ QPaug}                                & ANCE        & \textbf{+ QPaug}                                & Contriever  & \textbf{+ QPaug}                                \\ \cline{2-9} 
                         & NQ       & 35.1 / 37.9  & 35.5 / 40.3 & \textbf{40.7 / 44.2} & 35.5 / 39.8 & \textbf{40.9 / 45.4} & 35.2 / 37.9 & \textbf{41.6 / 44.6} \\
                         & 2wiki    & 19.7 / 27.1  & 20.2 / 30.2 & \textbf{24.0 / 35.3} & 20.0 / 26.0 & \textbf{24.6 / 35.7} & 20.3 / 26.4 & \textbf{24.5 / 35.5} \\
                         & Hotpot   & 27.4 / 35.3  & 28.3 / 38.3 & \textbf{38.0 / 44.9} & 29.0 / 34.8 & \textbf{37.3 / 44.4} & 28.2 / 33.7 & \textbf{38.4 / 45.1} \\ \cline{2-9} 
                         & Average  & 27.4 / 33.4  & 28.0 / 36.1 & \textbf{34.2 / 41.8}                           & 28.2 / 33.5 & \textbf{34.3 / 41.5}                           & 27.9 / 32.7 & \textbf{34.8 / 41.7}                           \\ \hline
\multirow{5}{*}{LLaMA-2-7b-chat}  & Datasets & No retrieval & SBERT       & \textbf{+ QPaug}      & ANCE        & \textbf{+ QPaug}      & Contriever  & \textbf{+ QPaug}      \\ \cline{2-9} 
                                & NQ       & 13.0 / 15.5  & 24.0 / 27.5 & \textbf{29.5 / 34.1} & 23.3 / 26.5 & \textbf{28.1 / 32.4} & 20.8 / 24.7 & \textbf{27.9 / 32.0} \\
                                & 2wiki    & 16.7 / 20.3  & 20.9 / 23.2 & \textbf{22.0 / 26.9}          & 20.6 / 23.0 & \textbf{22.5 / 25.8}          & 20.3 / 22.5 & \textbf{22.3 / 25.9} \\
                                & Hotpot   & 16.1 / 18.6  & 22.5 / 27.5 & \textbf{24.9 / 33.0} & 21.5 / 26.9 & \textbf{22.2 / 29.3} & 20.1 / 24.4 & \textbf{23.3 / 31.0} \\ \cline{2-9} 
                                & Average  & 15.3 / 18.1  & 22.5 / 26.1 & \textbf{25.5 / 31.3} & 21.8 / 25.5 & \textbf{24.3 / 29.1} & 20.4 / 23.9 & \textbf{24.5 / 29.6} \\ \hline
\end{tabular}
\end{adjustbox}
\caption{Performance (Rouge/F1 score) comparison between no-retreival, base RAG and QPaug across various retrievers and readers (LLMs). For example, the column labelled ``SBERT'' means a base RAG with SBERT as the retriever, and column labelled ``+QPaug'' on the right means we use QPaug as ``add-on'' to the RAG with SBERT retriever. GPT-4 is used for the question augmentation in QPaug.}
\label{tab:main_results}
\end{table*}

\noindent \textbf{Implementation Details.} Greedy decoding is used for LLM generations in all the experiments. We evaluate zero-shot QA tasks with retrieval where the retrieval is implemented based on \cite{thakur2021beir} with Faiss indexing~\cite{johnson2019billion}. We use Langchain\footnote{https://www.langchain.com/} for LLM inference with manually-designed prompt templates. Detailed prompt templates are provided in Appendix~\ref{appendix:prompt_templates}. 

\subsection{Main Results}

Table~\ref{tab:comparison_sure} shows the comparison between QPaug and other baseline methods, where Contriever is used as the retriever and GPT-3.5 as the LLM. Experimental results demonstrate that QPaug achieves a performance gain of 10.4\% on NQ, 8.9\% on 2wiki, and 34.2\% on HotpotQA over SuRE which is the current state-of-the-art. In addition, QPaug achieves average performance improvements of 22.6\% and 27.9\% over Chain-of-thoughts~\cite{kojima2022large} and Self-Verification~\cite{weng2023large} respectively. The results demonstrate that QPaug successfully can retrieve relevant context and generate factual knowledge supporting retrieved passages. 

Next, we examine the compatibility of QPaug with various types of retrievers and LLMs. Table~\ref{tab:main_results} summarizes the experimental results on ODQA benchmark datasets with three retrievers and LLMs. Each element indicates the combinations of retrievers (column) and LLMs~(row). We first apply QPaug to proprietary LLMs~\cite{achiam2023gpt}. GPT-4 with QPaug achieves performance gain on average from 14.8\% to 23.8\% in Rouge and from 20.1\% to 27.9\% in F1 score. In addition, GPT-3.5 achieves a similar performance gain on average of up to 24.5\% in Rouge and 27.5\% in F1 score when they are combined with QPaug. As an open-source LLM, LLaMA-2-7b-chat~\cite{touvron2023llama} achieves slightly lower average gains up to 20.1\% and 23.9\% in Rouge and F1 scores respectively. We observe that QPaug consistently improves the performances of ODQA irrespective of the types of LLMs and retrievers.

In particular, QPaug exhibits notable improvements in the multi-hop QA datasets of 2wiki and HotpotQA. GPT-4 with SBERT obtains performance gains of 2.1\% in Rouge and 17.4\% in F1 score on NQ; however, the same model achieves gains of 34.2\% / 37.9\% on 2wiki, and 14.5\% / 20.1\% on HotpotQA. The results show that QPaug achieves particularly large performance gains in multi-hop QA datasets, illustrating its effectiveness in solving complex questions. 

\section{Ablation study}
We conduct ablation study on two main components, i.e., question augmentation and passage generation, denoted by \emph{Qaug} and \emph{Pgen} in the following subsections.

\subsection{Question Augmentation (Qaug)}
\begin{table}[h!]
\begin{adjustbox}{width=.5\textwidth,center}
\begin{tabular}{llll}
\hline
Model / Dataset & NQ    & 2wiki & Hotpot \\
\hline
SBERT           & 63.41 & 27.90 & 47.47    \\
+Qaug by LLaMA-2-7b-chat & 63.74~(+0.33) & 29.21~(+1.31) & 52.07~(+4.60)    \\
+Qaug by LLaMA-2-70b-chat & 66.10~(+2.69) & 33.56~(+5.66) & 56.89~(+9.42)    \\
+Qaug by GPT-3.5        & 64.98~(+1.57) & 30.37~(+2.47) & 51.06~(+3.59)    \\
+Qaug by GPT-4          & 70.20~(+6.79) & 38.68~(+10.8) & 62.08~(+14.6)    \\
\hline
ANCE            & 60.43 & 22.57 & 38.81    \\
+Qaug by LLaMA-2-7b-chat & 62.91~(+2.48) & 24.04~(+1.47) & 43.88~(+5.07)    \\
+Qaug by LLaMA-2-70b-chat & 64.58~(+4.15) & 26.18~(+3.61) & 46.63~(+7.82)    \\
+Qaug by GPT-3.5        & 62.55~(+2.12) & 22.91~(+0.34) & 40.85~(+2.04)   \\
+Qaug by GPT-4          & 66.68~(+6.25) & 27.31~(+4.74) & 49.59~(+10.8)    \\

\hline
Contriever      & 52.48 & 21.26 & 40.54    \\
+Qaug by LLaMA-2-7b-chat & 58.10~(+5.62) & 25.00~(+3.74) & 47.78~(+7.42)    \\
+Qaug by LLaMA-2-70b-chat & 62.18~(+9.70) & 29.04~(+7.78) & 53.19~(+12.7)   \\
+Qaug by GPT-3.5        & 52.76~(+0.28) & 21.51~(+0.26) & 42.74~(+2.20)    \\
+Qaug by GPT-4          & 60.38~(+7.90) & 28.95~(+7.69) & 53.69~(+13.2)    \\
\hline
\end{tabular}
\end{adjustbox}
\caption{Answer Recall @10 of retrieved passages question augmentation.}
\label{tab:ablation_qaug_r10}
\end{table}

In this section, we investigate the effect of question augmentation (Qaug). Table~\ref{tab:ablation_qaug_r10} shows the change in the retrievers' performance measured in Recall@10 by adding only Qaug component. We experimented with various LLMs: GPT-4, GPT-3.5 and LLaMA-2. Table~\ref{tab:ablation_qaug_r10} demonstrates that Qaug boosts the retrieval performance by up to 30\% on average with GPT-4, where the improvements are greater than GPT-3.5. In addition, we experiment with both LLaMA-2 models of 7b and 70b sizes which achieve performance gains in all datasets where 70b size model obtains higher performance gains by a large margin. 

\begin{figure}[h!]
    \centering
    \includegraphics[width=.5\textwidth]{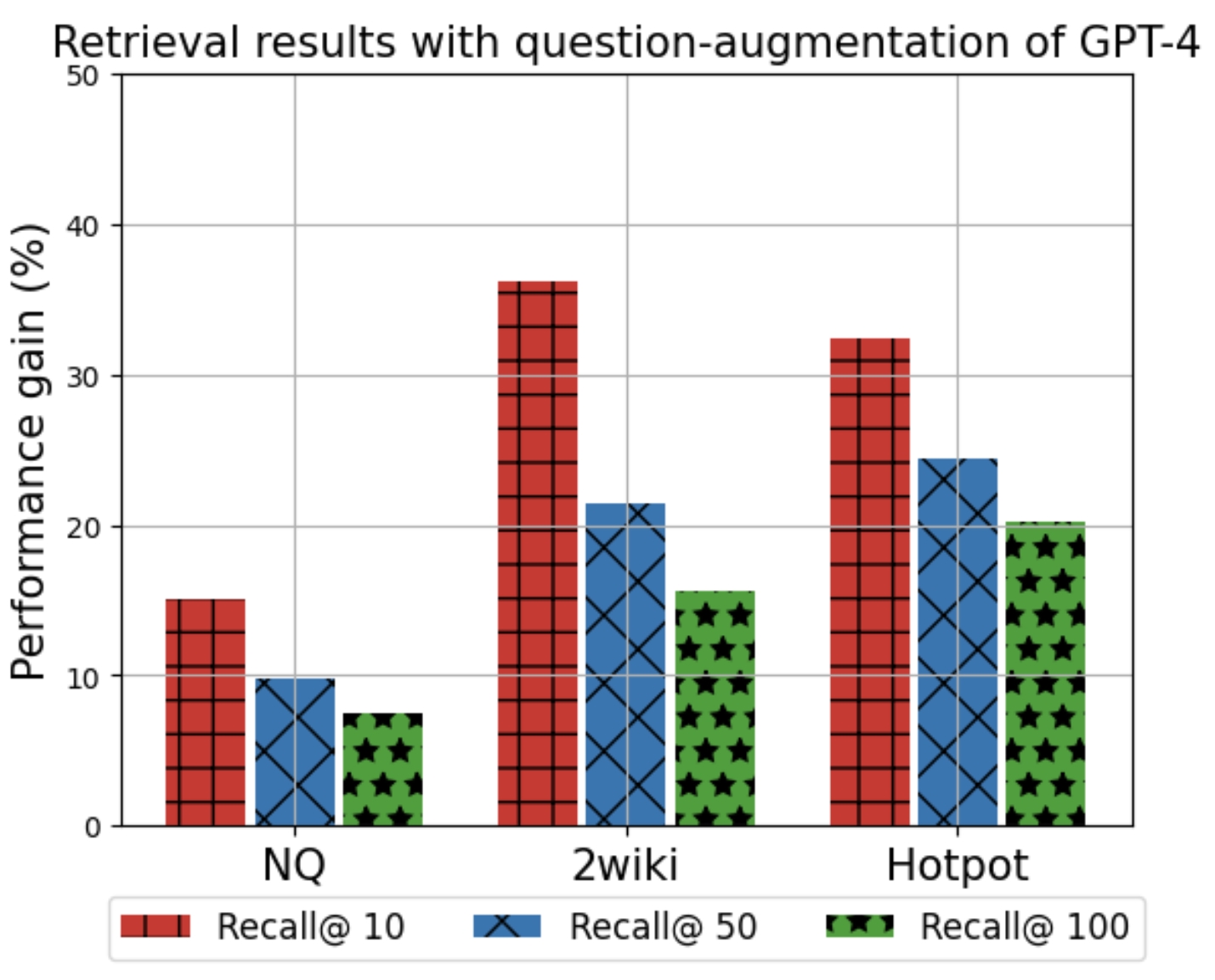}
    \caption{Performance gains of recall @ K with question augmentation of GPT-4. K is 10, 50, and 100. The base retriever is Contriever.}
    \label{fig:retrieval_results}
\end{figure}

Fig.~\ref{fig:retrieval_results} shows the performance gains achieved by Qaug with the varying number of retrieved passages $K$. We observe that the performance gain is more pronounced in $K=10$ than 50 or 100. This is because, if the number of passages to be retrieved is limited, the relevance and fine-granularity of queries composed by Qaug become more important for better retrieval. We also observe that Qaug improves the retrieval performances with a relatively large number of retrieved passages as well, i.e., with $K=50$ or 100.

\subsection{Passage Self-Generation (Pgen)}
In this section, we examine the effect of the passage self-generation (Pgen) component.
Table~\ref{tab:ablation_paug_gpt4}, \ref{tab:ablation_paug_gpt3.5} and \ref{tab:ablation_paug_llama7b} show the change in performance by adding the self-generated passage to retrieved passages, with GPT-4, GPT-3.5 and LLaMA-2-7b-chat respectively. 

Table~\ref{tab:ablation_paug_gpt4} shows that the addition of Pgen with GPT-4 improves performances in all datasets as compared to the baseline using only retrieved passages. In particular, the performance gain is significantly large at 35.1\% on 2wiki dataset for which the search/retrieval results tend to be relatively poor due to multi-hop questions. Pgen is particularly effective when the retrieval performance degrades. This demonstrates that the self-generated passage is able to complement the missing context in the retrieved passages. We observe similar trends with other LLMs, e.g., see Table~\ref{tab:ablation_paug_gpt3.5} and Table~\ref{tab:ablation_paug_llama7b} for the effects of Pgen with GPT-3.5 and LLaMA-2-7b-chat. The performance gains with GPT-3.5 are on average from 14.2\% to 25.9\%, and are on average from 11.6\% to 25.9\% with LLaMA-2-7b-chat. 

\begin{figure}[t!]
    \centering
    \includegraphics[width=.5\textwidth]{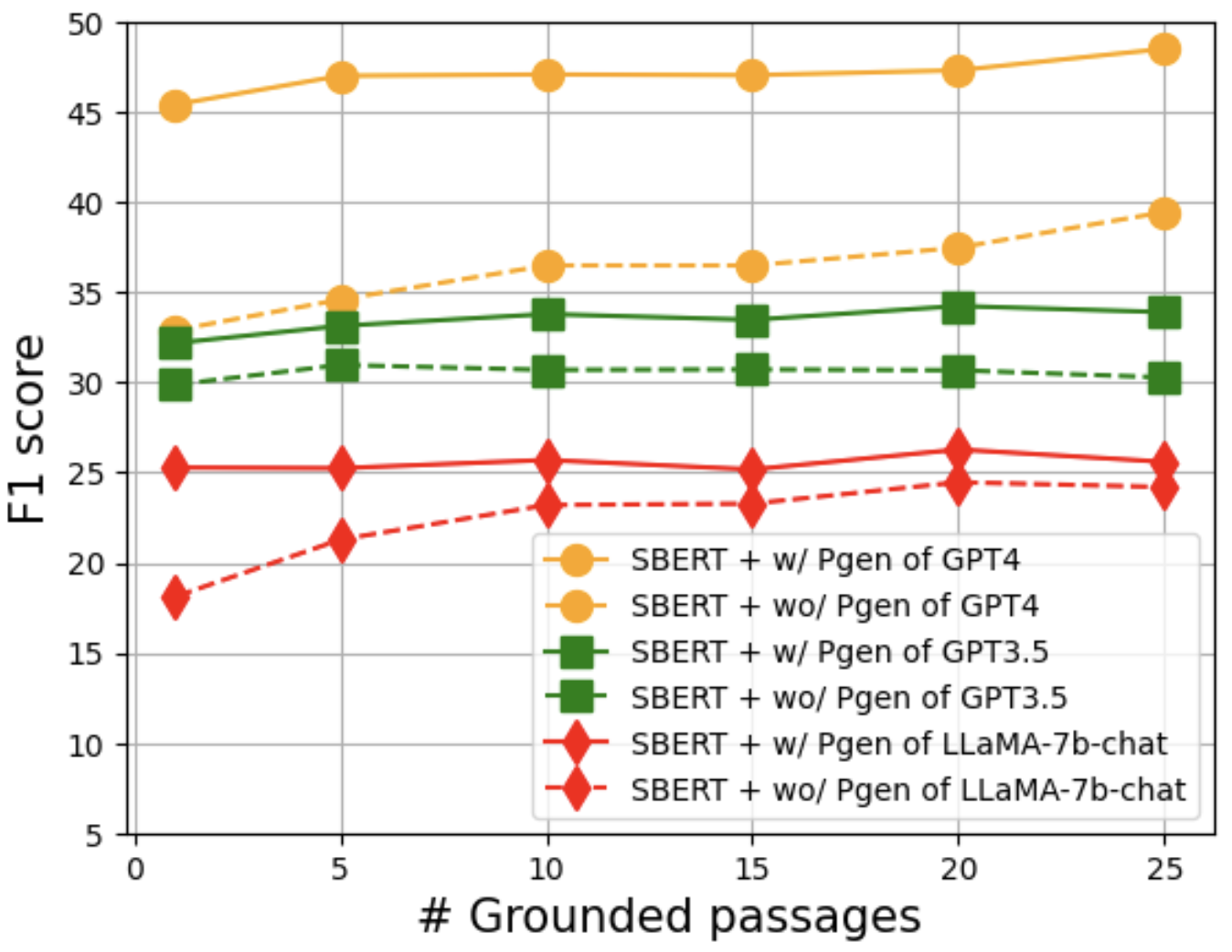}
    \caption{Comparison of F1-scores between RAG and QPaug per top-K grounded passages on 2wiki. Retriever is vanilla SBERT.}
    \label{fig:topk_passages}
\end{figure}

Finally, we examine the effect of Pgen by varying the number of retrieved passages $K$. Fig.~\ref{fig:topk_passages} shows the F1 scores per top-$K$ grounded passages. Similar to Fig.~\ref{fig:retrieval_results}, experimental results show that  Pgen achieves performance gains when the number of retrieved passages increases. Thus, LLM can extract factual knowledge even though there are a number of possibly irrelevant retrieved passages. Also, the results show that the passage generation method can be plugged into various LLMs since it illustrates the same trend for GPT-4, GPT-3.5, and LLaMA-2-7b-chat.

\begin{table}[t!]
\begin{tabular}{llll}
\hline
Model / Dataset & NQ   & 2wiki & Hotpot \\ \hline
SBERT           & 46.0 & 36.5  & 49.0   \\
+Pgen by GPT-4 & \textbf{47.6} & \textbf{47.8}  & \textbf{53.0} \\ \hline
ANCE            & 45.0 & 33.6  & 43.6   \\
+Pgen by GPT-4 & \textbf{47.8} & \textbf{45.6}  & \textbf{47.4}   \\ \hline
Contriever      & 43.7 & 32.6  & 43.2   \\
+Pgen by GPT-4 & \textbf{47.2} & \textbf{45.2}  & \textbf{47.7}   \\ \hline
\end{tabular}
\caption{F1 score between predictions and answers of GPT-4.}
\label{tab:ablation_paug_gpt4}
\end{table}

\begin{table}[t!]
\begin{tabular}{llll}
\hline
Model / Dataset       & NQ   & 2wiki & Hotpot \\
\hline
SBERT                 & 40.3 & 30.2  & 38.3     \\
+Pgen by GPT-3.5 & \textbf{44.0} & \textbf{34.2}  & \textbf{43.1}     \\
\hline
ANCE                  & 39.8 & 26.0  & 34.8     \\
+Pgen by GPT-3.5 & \textbf{44.2} & \textbf{35.2}  & \textbf{44.1}     \\
\hline
Contriever            & 37.9 & 26.4  & 33.7     \\
+Pgen by GPT-3.5 & \textbf{43.0} & \textbf{34.8}  & \textbf{44.6}  \\   
\hline
\end{tabular}
\caption{F1 score between predictions and answers of GPT-3.5.}
\label{tab:ablation_paug_gpt3.5}
\end{table}

\begin{table}[t!]
\begin{adjustbox}{width=.5\textwidth,center}
\begin{tabular}{llll}
\hline
Model / Dataset & NQ   & 2wiki & Hotpot \\ \hline
SBERT           & 27.5 & 23.2  & 27.5   \\
+Pgen by LLaMA-2-7b-chat & \textbf{31.2} & \textbf{26.2}  & \textbf{30.1} \\ \hline
ANCE            & 26.5 & 23.0  & 26.9   \\
+Pgen by LLaMA-2-7b-chat & \textbf{28.9} & \textbf{25.6}  & \textbf{29.0}   \\ \hline
Contriever      & 24.7 & 22.5  & 24.4   \\
+Pgen by LLaMA-2-7b-chat & \textbf{29.7} & \textbf{26.2}  & \textbf{29.0}   \\ \hline
\end{tabular}
\end{adjustbox}
\caption{F1 score between predictions and answers of LLaMA-2-7b-chat.}
\label{tab:ablation_paug_llama7b}
\end{table}

\section{Qualitative Analysis}
\paragraph{Effectiveness of Qaug.}

\begin{table*}[t!]
\begin{adjustbox}{width=1.\textwidth, center}
\begin{tabular}{|l|l|}
\hline
 \multicolumn{1}{|c|}{\textbf{Question}} & \multicolumn{1}{c|}{\textbf{Answer}} \\ \hline
 Who is the spouse of the director of film Eden And After? & {\color{blue}\textbf{Catherine Robbe-Grillet}}            \\ \hline \hline
\multicolumn{2}{|c|}{\textbf{Augmented Question by Qaug}}                                       \\ \hline
\multicolumn{2}{|p{\textwidth}|}{
\small
1. **Identify the Director**: First, we need to determine who directed "Eden And After". \newline
2. **Research the Director**: Once we have the name of the director, we can research more about his or her personal life, focusing on marital status to find out who their spouse is (or was, depending on current status).\newline
3. **Verify Information**: It's important to cross-check the information from multiple sources to ensure accuracy.\newline
...
} 
\\ \hline \hline
\multicolumn{2}{|c|}{\textbf{Retrieved Passage w/o Qaug}}                                       \\ \hline
\multicolumn{2}{|p{\textwidth}|}{
\small 
\textit{Top-1} \newline
and during ""I Dream of Jeannie"". Also covered are her three marriages to Michael Ansara (1958–1974), Charles Fegert (1977–1982) and Jon Eicholtz (1991–present) and her ""emotional breakdown"" following the death of her only child, ...}      \\ \hline \hline
\multicolumn{2}{|c|}{\textbf{Retrieved Passage w/ Qaug}} \\ \hline
\multicolumn{2}{|p{\textwidth}|}{
\small 
\textit{Top-1} \newline
... {\color{blue}\textbf{Catherine Robbe-Grillet}} (; ""née"" Rstakian; born 24 September 1930) is a French theatre and cinema actress, photographer, and writer of Armenian descent who has published sadomasochistic writings under ...
} 
\\ \hline
\end{tabular}
\end{adjustbox}
\caption{Qualitative comparison of retrieved passage with or without question augmentation~(Qaug) on 2wiki dataset. The question-augmentation method is GPT-4. The first passage results are obtained by searching using the original question, while the second passage results are obtained by searching using the augmented question. The complete search results and examples are provided in Appendix~\ref{appendix:QALM}.}
\label{tab:qalm-table}
\end{table*}

\begin{table*}[t!]
\begin{adjustbox}{width=1.\textwidth, center}
\begin{tabular}{|ll|}
\hline
\multicolumn{1}{|l|}{\textbf{Question}} & \textbf{Answer} \\ \hline
\multicolumn{1}{|l|}{Where was the place of death of the director of film Mole Men Against The Son Of Hercules?} & {\color{blue}\textbf{Rome}} \\ \hline \hline
\multicolumn{2}{|l|}{\textbf{Retrieved Passages}}                \\ \hline
\multicolumn{2}{|p{\textwidth}|}{
\small\textit{Top-1} \newline
... Against the Son of Hercules"" was released on 10 October 1961. Mole Men Against the Son of Hercules Mole Men Against
\newline
~\textit{Top-2} \newline
... poisoned by his own men (the ""sers felons"" Antipater and Divinuspater), as was another key figure of the work, Darius...
\newline
~\textit{Top-3} \newline
at Dubrava Film in Zagreb, Croatia and on location in Zagreb. ""The Fury of Hercules"" as released in Italy on 21 March 1962. The Fury of Hercules The Fury of Hercules () is a 1962 peplum film written and directed by Gianfranco Parolini...
}                                 \\ \hline \hline
\multicolumn{2}{|l|}{\textbf{Generated passage}}                    \\ \hline
\multicolumn{2}{|p{\textwidth}|}{
\small
The director of the film "Mole Men Against The Son Of Hercules" was Antonio Leonviola. He passed away on December 14, 1971, in ~{\color{blue}\textbf{Rome}}, Italy...
}                                 \\ \hline \hline
\end{tabular}
\end{adjustbox}
\caption{Qualitative analysis of generated passage (Pgen) compared to retrieved passages. The passage generation method is GPT-4. While retrieved passages do not include relevant context with the answer, generated passage includes the answer. The entire search results and additional examples are provided in Appendix~\ref{appendix:PALM}.}
\label{tab:palm-table}
\end{table*}

Table~\ref{tab:qalm-table} shows the comparison between the retrieval results using the original question versus the augmented question (Qaug) on 2wiki. The retrieval by Qaug successfully obtains relevant information containing the answer, while the retrieved passages from the original question do not contain relevant clues. As a result, the LLM outputs the correct answer, \textit{Catherine Robbe-Grillet} with Qaug, but outputs \textit{Not mentioned} with the original question
(for details, refer to Table~\ref{tab:qalm-2wiki} in Appendix~\ref{appendix:QALM}). Thus, Qaug enhances the retrieval of passages with relevant contexts, leading to significant improvement on the performance of ODQA.
\paragraph{Effectiveness of Pgen.}
Table~\ref{tab:palm-table} shows a qualitative analysis of passage generation (Pgen) using LLMs.
In this example, the Contriever fails to retrieve relevant passages. The top-1 passage provides a description of ~\textit{Mole Men Against the Son of Hercules}. The passage likely received the highest score due to the exact match of the phrase ~\textit{Mole Men Against the Son of Hercules} with the question.
The top-2 passage contains information about ~\textit{The Fury of Hercules}. It seems to have been retrieved based on its focus on ~\textit{Hercules}.
Lastly, the top-3 passage is also completely unrelated to the correct answer. It is due to getting a high score due to the presence of the words ~\textit{director} and ~\textit{Against the Son of Hercules}.
The rest of the passages in top-10 did not include content relevant to the correct answer. However, the LLM-generated passage accurately contained a short biography of ~\textit{Antonio Leonviola}, the director of ~\textit{Mole Men Against The Son Of Hercules}, which revealed a crucial hint on the place of death.
Results show that even though the retriever does not retrieve relevant context, Pgen complements the retrievers by their knowledge. Conversely, as mentioned earlier, there are cases where the LLM-generated passage did not provide relevant context to the question. Nevertheless, a correct answer was recovered in case the retrieved passages provided relevant information. Some examples are shown in Table~\ref{tab:pgen_none} and Table~\ref{tab:pgen_wrong} of Appendix \ref{appendix:PALM}.

\section{Conclusion}
In this paper, we propose a simple yet effective question and passage augmentation (QPaug) method via LLMs.
QPaug harmoniously combines parametric and non-parametric knowledge of LLMs by leveraging the massive knowledge of LLMs for designing queries for retrieval and for guiding answer extraction from passages. Experiments show that both proprietary and open-source LLMs successfully decompose questions into multiple sub-questions by the proposed question augmentation, which significantly improved the performances of passage retrieval. In addition, the self-generated passage was shown to achieve a large performance gain when it is augmented with the retrieved passages.

\section{Limitations}
Although we showed that the proposed question and passage augmentation method highly improves the performances on ODQA tasks, the detailed planning of questions and self-generation of passages rely much on LLMs' knowledge. However, the capability of LLMs are growing at an unprecedented rate, and we believe our work is timely in that it is reported that even LLM can generate synthetic data to train other language models and achieve good performance~\cite{li2023textbooks, gunasekar2023textbooks, benallal2024cosmopedia, abdin2024phi}, and such massive knowledge of LLMs should be more actively explored for many tasks, including RAG. 
In addition, although we explicitly instructed LLMs not to generate bogus passages, hallucinations still can occur. However, we believe that alignment techniques reducing LLMs' hallucinations are actively researched, which can further benefit our method in the future.

\section{acknowledgement}
This research was supported by the National Research Foundation of Korea (NRF) grant funded by the Korean government (MSIT) (No. 2022R1A5A1027646), the MSIT (Ministry of Science and ICT), Korea, under the ICT Creative Consilience program (IITP-2020-0-01819) supervised by the IITP (Institute for Information \& communications Technology Planning \& Evaluation).

\bibliography{custom}

\appendix

\clearpage
\onecolumn
\section{Appendix}
\subsection{Prompt templates}\label{appendix:prompt_templates}

\textbf{Prompts for answering the questions without passages.}
\begin{figure}[h!]
    \centering
    \fbox{
    \begin{minipage}{35em}
    Question: \{question\} Do not exceed 3 words.\\\\
    Answer:
    \end{minipage}
    }
    \caption{Prompt template design for retrieval-augmented answering questions.}
    \label{fig:prompt_base}
\end{figure}

\noindent\textbf{Prompts for answering the questions only using retrieved passages.}
\begin{figure}[h!]
    \centering
    \fbox{
    \begin{minipage}{35em}
    \{passages\}\\\\
    Question: \{question\} Do not exceed 3 words.\\\\
    Answer:
    \end{minipage}
    }
    \caption{Prompt template design for retrieval-augmented answering questions.}
    \label{fig:prompt_base}
\end{figure}

\noindent\textbf{Prompts for answering the questions using augmented passages.}
\begin{figure}[h!]
    \centering
    \fbox{
    \begin{minipage}{35em}
    \{passages\}\\\\
    Your knowledge: \{generated passage\}\\\\
    Question: \{question\} Do not exceed 3 words.\\\\
    Answer:
    \end{minipage}
    }
    \caption{Prompt template design for retrieval-augmented answering questions.}
    \label{fig:prompt_base}
\end{figure}

\noindent\textbf{Prompt template of passage generation.}
\begin{figure}[h!]
    \centering
    \fbox{
    \begin{minipage}{35em}
    Your job is to act as a subject matter expert. You will write a good-quality passage that can answer the question based on your factual knowledge. Do not write a passage if you don't know accurate information about the question.\\\\
    Now, let's start. After you write, please write [DONE] to indicate you are done. Write [NONE] if you cannot write a factual passage.\\
    Question: \{question\}\\
    Passage:
    \end{minipage}
    }
    \caption{Prompt template design for generating knowledgeable passages.}
    \label{fig:prompt_palm}
\end{figure}

\subsection{Additional Qualitative Results}
\label{appendix:Qualitative}
\subsubsection{Retrieval Results from Original Question and Augmented Question}
\label{appendix:QALM}
\begin{table}
\begin{adjustbox}{width=\textwidth, center}
\begin{tabular}{|l|l|}
\hline
 \multicolumn{1}{|c|}{\textbf{Question}} & \multicolumn{1}{c|}{\textbf{Answer}} \\ \hline
 who was the buccaneers qb when they won the superbowl & {\color{blue}\textbf{Brad Johnson}} \\ \hline \hline
\multicolumn{2}{|c|}{\textbf{Retrieved Passages (w/o Qaug)}}                                       \\ \hline
\multicolumn{2}{|p{\textwidth}|}{
\scriptsize
\textit{Top-1} \newline
in as QB due to Mark Sanchez suffering an apparent shoulder injury. His first game didn\'t start off on the right foot, though. He fumbled after taking a sack from Buccaneers MLB Mason Foster, and threw an interception to Buccaneers linebacker Lavonte David. Smith eventually rebounded by throwing his first TD pass to Kellen Winslow Jr. to make the game close by halftime. The second half was a defensive struggle between both teams that wound up in both teams scoring in field goals. The biggest moment of the game came when Geno Smith scrambled out-of-bounds and Lavonte David (the same
\newline\textit{Top-2} \newline
later became head coach of the Seattle Seahawks where they played in Super Bowl XL after the 2005 season. However, the Seahawks lost to the Pittsburgh Steelers. One of Holmgren\'s former assistants, Jon Gruden, has had reasonable success running the West Coast offense in his own right. He started his head coaching career with the Oakland Raiders, leading them from 1998-2001, and turned the Raiders into a strong playoff contender. Gruden then went on to become head coach of the Tampa Bay Buccaneers, winning Super Bowl XXXVII after the 2002 season. Gruden coached the Buccaneers from 2002-2008. After several years
\newline\textit{Top-3} \newline
\textbf{Johnson} again had another great season in Tampa Bay and won a Super Bowl the following year. In 2001, {\color{blue}\textbf{Brad Johnson}} was pursued by the Baltimore Ravens in the off season (coached by Brian Billick, former offensive coordinator for the Vikings) but he spurned them to join the Bucs. \textbf{Johnson} was reunited with former Vikings assistant-coach Tony Dungy for his first season with the Tampa Bay Buccaneers. That year, he broke Tampa Bay team records for passing yards with 3,406, completions with 340, and attempts with 540. In the 2002 season he led the Buccaneers to their first ever Super
\newline\textit{Top-4} \newline
QB Richard Todd in 1976. In his tenure with the Jets, he threw for more interceptions than he did touchdowns. In the \'81 season, the Jets played vs the Miami Dolphins in the AFC Championship Game. Todd threw for 5 interceptions and the Jets lost the game. A year later, Todd would be traded to the New Orleans Saints. The most recent bust, Dee Milliner, was drafted by the team in 2013. Milliner played his college career at the University of Alabama and had high expectations after being drafted. Lasting just 3 years with the team, Milliner\'s career was plagued
\newline\textit{Top-5} \newline
when he served as head coach of the San Diego Chargers from 2007-2012. His Chargers teams showcased the talents of QB Philip Rivers, RB LaDainian Tomlinson and TE Antonio Gates. The St. Louis Rams ran the Coryell system successfully under coordinator and then head coach Mike Martz. Martz served as St. Louis\'s offensive coordinator under head coach Dick Vermeil in the 1999 season, when the Rams won Super Bowl XXXIV. Martz then served as the Rams head coach 2000-2005. His teams were anchored offensively by QB Kurt Warner and RB Marshall Faulk, both of whom are Hall of Famers. Earlier
}      \\ \hline 
\multicolumn{2}{|l|}{\textbf{Answer prediction w/o Qaug}: {\color{blue}\textbf{Brad Johnson}}}  \\ \hline \hline
\multicolumn{2}{|c|}{\textbf{Retrieved Passages (w/ Qaug)}} \\ \hline
\multicolumn{2}{|p{\textwidth}|}{
\scriptsize
\textit{Top-1} \newline
{\color{blue}\textbf{Johnson}} again had another great season in Tampa Bay and won a Super Bowl the following year. In 2001, {\color{blue}\textbf{Brad Johnson}} was pursued by the Baltimore Ravens in the off season (coached by Brian Billick, former offensive coordinator for the Vikings) but he spurned them to join the Bucs. \textbf{Johnson} was reunited with former Vikings assistant-coach Tony Dungy for his first season with the Tampa Bay Buccaneers. That year, he broke Tampa Bay team records for passing yards with 3,406, completions with 340, and attempts with 540. In the 2002 season he led the Buccaneers to their first ever Super
\newline \textit{Top-2} \newline
XXXVII in 2003 under coach Jon Gruden. Tampa has hosted four Super Bowls: Super Bowl XVIII (1984), Super Bowl XXV (1991), Super Bowl XXXV (2001), and Super Bowl XLIII (2009). The first two events were held at Tampa Stadium, and the other two at Raymond James Stadium. Tampa will be the host for Super Bowl LV in 2021. Originally the Pittsburgh Gladiators and a charter member of the Arena Football League (AFL), the Tampa Bay Storm relocated from Pittsburgh in 1991 and won ArenaBowl V that year. They later won 4 more ArenaBowls (VII, IX, X, and XVII, and also
\newline \textit{Top-3} \newline
Super Bowl ring following the Buccaneers\' victory in Super Bowl XXXVII. He signed with the Arena Football League\'s Orlando Predators in 2004 and guided the team to a 9-5 record and the playoffs, despite suffering another knee injury and missing two and a half games. He was then signed by the Indianapolis Colts in 2004, reuniting with former Buccaneers coach Tony Dungy, but only saw limited action in one game before being released during the season. He returned to the Orlando Predators where he was the starting quarterback through the 2006 season. He has a 32-15 record as the Predators\'
\newline \textit{Top-4} \newline
Play of season QB Rating = Passer rating W/L Record = Super Bowl/Postseason Won/Loss Record After his retirement from professional football as a player, Aikman joined Fox\'s NFC telecasts as a color commentator for the 2001 season. A year later, he was named to the network\'s lead announcing crew, teaming with Joe Buck and (from 2002–2004) Cris Collinsworth. Aikman received an Emmy Award nomination for his television work in 2004 and has helped broadcast five Super Bowls (XXXIX, XLII, XLV, XLVIII, and LI) to date. It was revealed in 2016 that in 2004, Aikman nearly came out of retirement to
\newline \textit{Top-5} \newline
championship games (XVII, XXV, XXVIII, XXXIV, and XXXVI). It was also the last Super Bowl played in the month of January. This was the first Super Bowl in which the league\'s number one-ranked offense (Raiders) faced the league\'s number one-ranked defense (Buccaneers). The game sometimes is referred to as the ""Gruden Bowl"", because the primary storyline surrounding the game revolved around Jon Gruden. Gruden was the Raiders\' head coach from 1998 to 2001, and then became the Buccaneers coach in 2002. Tampa Bay, ""Gruden\'s ""new"" team"", made their first Super Bowl appearance in team history after posting a regular season
}
\\ \hline
\multicolumn{2}{|l|}{\textbf{Answer prediction w/ Qaug}: {\color{blue}\textbf{Brad Johnson}}}  \\ \hline 
\end{tabular}
\end{adjustbox}
\caption{Result of retrieved passages from the original question and augmented question on NQ.}
\label{tab:qalm-nq}
\end{table}

\begin{table*}[t!]
\begin{adjustbox}{width=1.\textwidth, center}
\begin{tabular}{|l|l|}
\hline
 \multicolumn{1}{|c|}{\textbf{Question}} & \multicolumn{1}{c|}{\textbf{Answer}} \\ \hline
 Who is the spouse of the director of film Eden And After? & {\color{blue}\textbf{Catherine Robbe-Grillet}}           \\ \hline \hline
\multicolumn{2}{|c|}{\textbf{Retrieved Passage (w/o Qaug)}}                                       \\ \hline
\multicolumn{2}{|p{\textwidth}|}{
\scriptsize 
\textit{Top-1} \newline
and during ""I Dream of Jeannie"". Also covered are her three marriages to Michael Ansara (1958–1974), Charles Fegert (1977–1982) and Jon Eicholtz (1991–present) and her ""emotional breakdown"" following the death of her only child, Matthew Ansara. On November 17, 1988, \textit{Eden} received a star on the Hollywood Walk of Fame for her contributions to television. In 1990, the University of West Los Angeles School of Law granted \textit{Eden} an honorary Doctor of Laws degree. Barbara Eden Barbara \textit{Eden} (born Barbara Jean Morehead, August 23, 1931) is an American film, stage, and television actress, and singer, best known for her starring
\newline\textit{Top-2} \newline
to his lawyer, the filmmaker informed French tax authorities in the month preceding the release of the papers about his offshore holdings. Mareva Grabowski is listed in the Paradise Papers. She is the wife of Kyriakos Mitsotakis, who is leader of the opposition and president of the New Democracy political party. She owns 50\% of an offshore company, Eternia Capital Management in the Cayman Islands. This match is verified in Appleby and on listed in Cayman records on 30 March 2010. In total, 2.829 Greek names are listed in the papers. U2 lead singer Bono is listed in the papers
\newline\textit{Top-3} \newline
Lauren and Michael are having a hard time not playing those guardian roles. I like that \textit{Eden} wants to be her own person and I really enjoy bringing that feistiness to her. Since being in a relationship with Noah, living in Paris and then going to rehab for an eating disorder, she has to be a different person."" River Baldwin, ex-husband of Gloria Bardwell and absentee father of Michael Baldwin, was first seen onscreen in 2008, and was rumored to have a daughter in which he raised on an Ashram. River\'s daughter arrived in town, operating under the alias \textit{Eden}
\newline\textit{Top-4} \newline
with Sergio Leone`s restored Once Upon a Time in the West, in the presence of Claudia Cardinale. Lívia Gyarmathy is one of the greatest figures of Hungarian cinema. Even her first film, Do You Know Sunday-Monday? (Ismeri szandi-mandit?), was well-perceived both by the audience and the critics. The film, starring local legends Ila Schütz, Margit Dajka and Manyi Kiss is still a classic and repeatedly screened. The effects of the political changes of 1989 are pictured in Rapture of Deceit, starring Rita Tushingham. Besides her fine feature films she is also well known for her documentaries, like the touching The
\newline\textit{Top-5} \newline
she went on to answer her own questions through film: “The war lingers in my head, and I always search for it traces. So I wanted to ask my questions on the screen”. She began her socio-political documentary work around the year 2000. When asked about her objectivity in “Sleepless Nights” Eliane responded: “In the field of cinema, objectivity is a big lie, I tried to be as objective as possible, that is why I did not make a film about the Lebanese civil war, rather I made a film about two people who lived through the war. I began\newline
...
}      \\ \hline 
\multicolumn{2}{|l|}{\textbf{Answer prediction of w/o Qaug}: {\color{red}Not mentioned}}     \\ \hline \hline
\multicolumn{2}{|c|}{\textbf{Retrieved Passage (w/ Qaug)}} \\ \hline
\multicolumn{2}{|p{\textwidth}|}{
\scriptsize
\textit{Top-1} \newline
writer and filmmaker \textbf{Alain Robbe-Grillet} in Paris on 23 October 1957; he died in February 2008. In 2014, she was the subject of a documentary film entitled ""The Ceremony"", which examines her life as a member of the BDSM (sadomasochistic) community. {\color{blue}\textbf{Catherine Robbe-Grillet Catherine Robbe-Grillet}} (; ""née"" Rstakian; born 24 September 1930) is a French theatre and cinema actress, photographer, and writer of Armenian descent who has published sadomasochistic writings under the pseudonyms Jean de Berg and Jeanne de Berg. She was born in Paris, where she attended secondary school and high school. ""L\'Image"", a sadomasochistic novel published in 1956 
\newline \textit{Top-2} \newline
a decade before the appearance of his next feature film, ""La belle captive"" (""The Beautiful Captive"") (1983), but \textbf{Alain Robbe-Grillet} was fortunate enough to enlist the services of Henri Alekan as cinematographer, the visionary master of cinematography for the films of Jean Cocteau. Subsequently, more than a decade passed before \textbf{Alain Robbe-Grillet} got behind the lens again, this time filming a mystery thriller on a small Greek island with Fred Ward starring as the confused Frank in ""Un bruit qui rend fou"". \textbf{Robbe-Grillet} (""A Maddening Noise"", aka: ""The Blue Villa"") (1995). Before his death in 2008 \textbf{Robbe-Grillet} was to direct
\newline \textit{Top-3} \newline
1982); references to works are often accompanied by their ""H"" (for Hitchcock) number. The following lists show selected pieces, not his entire production in each genre. Marc-Antoine Charpentier Marc-Antoine Charpentier (; 1643 – 24 February 1704) was a French composer of the Baroque era. Exceptionally prolific and versatile, Charpentier produced compositions of the highest quality in several genres. His mastery in writing sacred vocal music, above all, was recognized and hailed by his contemporaries. Any family relationship between him and Gustave Charpentier, the late-nineteenth and early-twentieth century French opera composer, is highly unlikely. Charpentier was born in or near Paris
\newline \textit{Top-4} \newline
of the male protagonist, who becomes the narrative voice, instead of the female protagonist, Violette (Catherine Jourdan). N. a pris les dés... N. a pris les dés... (, French for ""N. has taken the dice..."") is a 1971 French experimental independent underground drama art film directed by \textbf{Alain Robbe-Grillet}. \textbf{Alain Robbe-Grillet} had signed for a production of two separate films from the same shot, with different editings of the same scenes, so to create two totally different plots: the first was ""Eden and After"", the second ""N. a pris les dés..."", whose title is indeed an anagram of the other
\newline \textit{Top-5} \newline
""Captive"". The book is referred to as a ""roman"" (novel) and is illustrated with 77 paintings by Magritte interspersed with discourse written by \textbf{Robbe-Grillet}. The eponymous film ""La Belle captive"", written and directed by Robbe-Grillet, was released in 1983. \textbf{Alain Robbe-Grillet Alain Robbe-Grillet} (; 18 August 1922 – 18 February 2008) was a French writer and filmmaker. He was one of the figures most associated with the ""Nouveau Roman"" (new novel) trend of the 1960s, along with Nathalie Sarraute, Michel Butor and Claude Simon. \textbf{Alain Robbe-Grillet} was elected a member of the Académie française on 25 March 2004, succeeding Maurice \newline
... \newline
\newline \textit{Top-10} \newline
... \textbf{Alain Robbe-Grillet Alain Robbe-Grillet} (; 18 August 1922 – 18 February 2008) was a French writer and filmmaker. He was one of the figures most associated with the ""Nouveau Roman"" (new novel) trend of the 1960s, along with Nathalie Sarraute, Michel Butor and Claude Simon. \textbf{Alain Robbe-Grillet} was elected a member of the Académie française on 25 March 2004, succeeding Maurice Rheims at seat No. 32. \textbf{He was married} to {\color{blue}\textbf{Catherine Robbe-Grillet (née Rstakian)}}. \textbf{Alain Robbe-Grillet} was born in Brest (Finistère, France) to a family of engineers and scientists. He was trained as an agricultural engineer. During the years 1943
}
\\ \hline 
\multicolumn{2}{|l|}{\textbf{Answer prediction of w/ Qaug}: {\color{blue}\textbf{Catherine Robbe-Grillet}}}          \\ \hline 
\end{tabular}
\end{adjustbox}
\caption{Result of retrieved passages from the original question and augmented question 2wiki dataset.}
\label{tab:qalm-2wiki}
\end{table*}

\begin{table}
\begin{adjustbox}{width=\textwidth, center}
\begin{tabular}{|l|l|}
\hline
 \multicolumn{1}{|c|}{\textbf{Question}} & \multicolumn{1}{c|}{\textbf{Answer}} \\ \hline
 \multirow{2}{*}{\shortstack[l]{
 Where is the stadium at which 1964 Georgia Tech Yellow Jackets \\
football team played their home game located?}} & \multirow{2}{*}{{\color{blue}\textbf{North Avenue at Techwood Drive}}} \\ & \\ \hline \hline
\multicolumn{2}{|c|}{\textbf{Retrieved Passage (w/o Qaug)}}                                       \\ \hline
\multicolumn{2}{|p{\textwidth}|}{
\scriptsize
\textit{Top-1} \newline
field fence and wall could be subject to damage from long home runs. The O\'Keefe lot and others nearby (Architecture and Van Leer Electrical Engineering lots) are no longer available due to construction or other campus projects. Consult The Georgia Tech Athletic Association, Rusty C, or Beesball.com website links provided in the ""External links"" section below for maps and more detailed information on recommended parking. Russ Chandler Stadium Russ Chandler Stadium is a college baseball stadium in Atlanta, Georgia. It has been the home field of the Georgia Tech Yellow Jackets college baseball team since 1930. The current stadium opened
\newline\textit{Top-2} \newline
Tech blockers 78 yards down the gridiron into the endzone as time expired, giving Georgia Tech its third win of the 2015 season, and third Top 10 win under Paul Johnson\'s guidance. {\color{orange}\textbf{Bobby Dodd Stadium Bobby Dodd Stadium}} at Historic Grant Field is the football stadium located at the corner of \textbf{North Avenue at Techwood Drive} on the campus of the Georgia Institute of Technology in Atlanta. It has been home to the Georgia Tech Yellow Jackets football team, often referred to as the ""Ramblin\' Wreck"", in rudimentary form since 1905 and as a complete stadium since 1913. The team
\newline\textit{Top-3} \newline
the Yellow Jackets home field. A new facility being constructed adjacent to historic O\'Keefe High School was completed February 2009, supplanting the twenty-year-old Glenn Field. The new facility will hold 1,000 spectators and make it possible for the Jackets to host postseason games with the presence of stadium lights. Glenn Field Glenn Field is the former home field for the Georgia Tech Yellow Jackets fast pitch softball team located in Midtown Atlanta, Georgia. Glenn Field was built in 1987 and has a capacity of 500 spectators. Glenn Field is one of the two Tech sports facilities off campus being located
\newline\textit{Top-4} \newline
in Atlanta and Georgia in Athens. ""We simply could not stop them."" - Bobby Bowden Pregame Line: Virginia Tech –3.5 GT vs. Virginia Tech Hokies Josh Nesbitt scored three touchdowns and Georgia Tech ran for 306 yards to upset \#4 Virginia Tech on a cold, memorable night in Atlanta. For the second straight week, head coach Paul Johnson led the \#19 Yellow Jacket squad to a victory, earning a win that ended a 17-game losing skid to top five opponents played at \textbf{Bobby Dodd Stadium} since 1962. Jason Peters batted down and intercepted Tyrod Taylor in the 1st quarter. Josh
\newline\textit{Top-5} \newline
where they still are today. Piedmont Park served the Atlanta Crackers, the city\'s original professional baseball team, before they moved to a stadium on Ponce de Leon Avenue in 1904. The two baseball teams have met 345 times since 1898. Georgia Tech has 148 wins, Georgia has 195 wins, and there are 2 ties in the series. Three baseball games are played between the two institutions every year. Two of the three games are played at the respective colleges\' baseball stadiums while the finale is played at SunTrust Park, home of the Atlanta Braves. The 2004 Georgia Tech vs. Georgia
}      \\ \hline 
\multicolumn{2}{|l|}{\textbf{Answer prediction w/o Qaug}: {\color{orange}Bobby Dodd Stadium}}  \\ \hline \hline
\multicolumn{2}{|c|}{\textbf{Retrieved Passage (w/ Qaug)}} \\ \hline
\multicolumn{2}{|p{\textwidth}|}{
\scriptsize
\textit{Top-1} \newline
Tech blockers 78 yards down the gridiron into the endzone as time expired, giving Georgia Tech its third win of the 2015 season, and third Top 10 win under Paul Johnson\'s guidance. {\color{orange}\textbf{Bobby Dodd Stadium Bobby Dodd Stadium}} at Historic Grant Field is the football stadium located at the corner of {\color{blue}\textbf{North Avenue at Techwood Drive}} on the campus of the Georgia Institute of Technology in Atlanta. It has been home to the Georgia Tech Yellow Jackets football team, often referred to as the ""Ramblin\' Wreck"", in rudimentary form since 1905 and as a complete stadium since 1913. The team
\newline \textit{Top-2} \newline
of the 2017 MLS season at {\color{orange}\textbf{Bobby Dodd Stadium}} until Mercedes-Benz Stadium was ready in mid-2017. The soccer club paid the university and the athletic association over \$1 million for the usage of {\color{orange}\textbf{Bobby Dodd Stadium}}. October 7, 1916: Georgia Tech 222, Cumberland College 0<br>In the most lopsided game in American football history, Georgia Tech, under coach John Heisman, defeated Cumberland College, 222–0. It has been said that Coach Heisman was repaying the Bulldogs for a 22–0 baseball defeat the previous year in which Cumberland had allegedly used professional players to ensure victory or conversely that he was showing how
\newline \textit{Top-3} \newline
in Atlanta and Georgia in Athens. ""We simply could not stop them."" - Bobby Bowden Pregame Line: Virginia Tech –3.5 GT vs. Virginia Tech Hokies Josh Nesbitt scored three touchdowns and Georgia Tech ran for 306 yards to upset \#4 Virginia Tech on a cold, memorable night in Atlanta. For the second straight week, head coach Paul Johnson led the \#19 Yellow Jacket squad to a victory, earning a win that ended a 17-game losing skid to top five opponents played at {\color{orange}\textbf{Bobby Dodd Stadium}} since 1962. Jason Peters batted down and intercepted Tyrod Taylor in the 1st quarter. Josh
\newline \textit{Top-4} \newline
field fence and wall could be subject to damage from long home runs. The O\'Keefe lot and others nearby (Architecture and Van Leer Electrical Engineering lots) are no longer available due to construction or other campus projects. Consult The Georgia Tech Athletic Association, Rusty C, or Beesball.com website links provided in the ""External links"" section below for maps and more detailed information on recommended parking. Russ Chandler Stadium Russ Chandler Stadium is a college baseball stadium in Atlanta, Georgia. It has been the home field of the Georgia Tech Yellow Jackets college baseball team since 1930. The current stadium opened
\newline \textit{Top-5} \newline
{\color{orange}\textbf{Bobby Dodd Stadium Bobby Dodd Stadium}} at Historic Grant Field is the football stadium located at the corner of {\color{blue}\textbf{North Avenue at Techwood Drive}} on the campus of the Georgia Institute of Technology in Atlanta. It has been home to the Georgia Tech Yellow Jackets football team, often referred to as the ""Ramblin\' Wreck"", in rudimentary form since 1905 and as a complete stadium since 1913. The team participates in the NCAA Division I Football Bowl Subdivision as a member of the Atlantic Coast Conference. It is the oldest stadium in the FBS and has been the site of more
}
\\ \hline
\multicolumn{2}{|l|}{\textbf{Answer prediction w/ Qaug}: {\color{orange}Bobby Dodd Stadium}}  \\ \hline 
\end{tabular}
\end{adjustbox}
\caption{Result of retrieved passages from the original question and augmented question on HotpotQA.}
\label{tab:qalm-hotpotqa}
\end{table}

Table~\ref{tab:qalm-2wiki}, Table~\ref{tab:qalm-hotpotqa}, and \ref{tab:qalm-nq} show search results from the question and augmented question on NQ, 2wiki, and Hotpot QA datasets.

Table~\ref{tab:qalm-nq} demonstrates that the retrieval from original question retrieves the passage containing the answer in the top-3 results, while the augmented question retrieves it at the top-1. This example demonstrates that using the augmented question improves the search process, leading to higher-quality search results. Also, when passages containing clues about the correct answer are placed at the beginning of the prompt, it can be more helpful in generating the answer, as observed by~\cite{lostmid}.

In the example of the 2wiki dataset (Table~\ref{tab:qalm-2wiki}), the question is asking about the spouse of the director of the film \textit{Enden And After}. The search results using only the original question mainly focus on the word \textit{Eden}, and we observe that the overall content of the retrieved passages is disconnected from the correct answer. Based on these search results, the output answer is \textit{Not mentioned}. By contrast, augmented question retrieves passages containing \textit{director of film Eden And After} throughout the searched passages. The top-1 document includes the correct answer, \textbf{Catherine Robbe-Grillet}. Moreover, the top-10 passages contain clear sentences that indicate the relationship between the director and their spouse, providing the necessary information to answer the question.

The last example, Table~\ref{tab:qalm-hotpotqa} shows that both retrieved passages from the original question and retrieved passages from the augmented question predict the same answer, \textbf{Bobby Dodd Stadium}. The address of this output is \textit{177 North Avenue NW, Atlanta, GA 30332, USA}, which is the same location as the correct answer. Consequently, we observe that even if the LLM's answer does not exactly match the correct answer, it can respond with a semantically equivalent entity. Comparing the search results, the augmented question retrieves passages containing the correct answer as more relevant passages compared to retrieval from original questions. It seems that the augmented question can find one more relevant passage, providing sufficient information for grounding the LLM.

\subsubsection{Qualitative Results for generated passage}
\label{appendix:PALM}

\begin{table*}[t!]
\begin{adjustbox}{width=1.\textwidth, center}
\begin{tabular}{|ll|}
\hline
\multicolumn{1}{|l|}{\textbf{Question}} & \textbf{Answer} \\ \hline
\multicolumn{1}{|l|}{Where was the place of death of the director of film Mole Men Against The Son Of Hercules?} & {\color{blue}\textbf{Rome}} \\ \hline \hline
\multicolumn{2}{|l|}{\textbf{Retrieved passages}}                \\ \hline
\multicolumn{2}{|p{\textwidth}|}{
\scriptsize
\textit{Top-1} \newline
Trapped in the caverns, the now-freed slaves break out onto the surface, thanks to Maciste's strength. Halis Mosab, who has become aware that she was captured as a child and somehow became queen of the 'mole men', is overjoyed when she comes to the surface and finds that she can live in the sunlight. But she accidentally falls off a cliff and into the waters of the Sacred Waterfall. The fate of her people is not revealed. ""Mole Men Against the Son of Hercules"" was released on 10 October 1961. Mole Men Against the Son of Hercules Mole Men Against
\newline
~\textit{Top-2} \newline
Alexander will not realize this goal: poisoned by his own men (the ""sers felons"" Antipater and Divinuspater), as was another key figure of the work, Darius. The cause of the death of the hero is shroudded in ambiguity: prophetic in nature, it poses the problem of divine retribution, patricide and political mistakes. Was Alexander killed because of his desire to elevate himself to the gods\' level? Was he killed by the magician Nectanabo, who is his father in the Greek and Roman tradition, and who also presided over his birth (Alexander kills him in a spite of rage)? Or was
\newline
~\textit{Top-3} \newline
at Dubrava Film in Zagreb, Croatia and on location in Zagreb. ""The Fury of Hercules"" as released in Italy on 21 March 1962. The Fury of Hercules The Fury of Hercules () is a 1962 peplum film written and directed by Gianfranco Parolini. Hercules is reached by the slave Daria, who informes him that his country has fallen into the hands of the tyrant Meniste. Hercules follows Daria in his homeland, where he discovers that Meniste enslaves the citizens of the city, and that a group of rebels are trying an insurrection. Meniste fears the power of Hercules, however he
\newline
~\textit{Top-4} \newline
at Dubrava Film in Zagreb, Croatia and on location in Zagreb. ""The Fury of Hercules"" as released in Italy on 21 March 1962. The Fury of Hercules The Fury of Hercules () is a 1962 peplum film written and directed by Gianfranco Parolini. Hercules is reached by the slave Daria, who informes him that his country has fallen into the hands of the tyrant Meniste. Hercules follows Daria in his homeland, where he discovers that Meniste enslaves the citizens of the city, and that a group of rebels are trying an insurrection. Meniste fears the power of Hercules, however he
\newline
~\textit{Top-5} \newline
his middle initial ""M"" as the name Anthony Dawson was used by the English actor. He was the only Italian director who worked directly for American production companies like MGM, United Artists, 20th Century Fox, Columbia Pictures, etc. with films like: ""Yor, the Hunter from the Future"", ""Take a Hard Ride"", ""Killer Fish"", etc. Margheriti said his action/adventure/war films were his favorites, and his least favorite movies were the sword-and-sandal ""peplum"" films he made in the early 1960s (such as ""Devil of the Desert Against the Son of Hercules"" and ""Giants of Rome""). For years, director Paul Morrissey disputed Margheriti\'s\newline
...
}                                 \\ \hline \hline
\multicolumn{2}{|l|}{\textbf{Generated passage}}                    \\ \hline
\multicolumn{2}{|p{\textwidth}|}{
\small
The director of the film "Mole Men Against The Son Of Hercules" was Antonio Leonviola. He passed away on December 14, 1971, in {\color{blue}\textbf{Rome}}, Italy. Antonio Leonviola was a prominent Italian director known for his work in the peplum genre, including films like "Mole Men Against The Son Of Hercules." His contributions to Italian cinema have left a lasting impact on the industry.
}                                 \\ \hline \hline
\multicolumn{2}{|l|}{\textbf{Answer Prediction}}                     \\ \hline
\multicolumn{2}{|p{\textwidth}|}{
\small
{\color{blue}\textbf{Rome}}}\\
\hline
\end{tabular}
\end{adjustbox}
\caption{Qualitative result of retrieved passages with Pgen (Ours) compared to Base RAG for 2wiki.}
\label{tab:palm-1}
\end{table*}

\begin{table*}[t!]
\begin{adjustbox}{width=1.\textwidth, center}
\begin{tabular}{|l|l|}
\hline
\multicolumn{1}{|l|}{\textbf{Question}} & \textbf{Answer} \\ \hline
Where did the director of film Playing It Wild die? & \textbf{\color{blue}Hollywood\color{black}} \\ \hline \hline
\multicolumn{2}{|l|}{\textbf{Retrieved passages}}                \\ \hline
\multicolumn{2}{|p{\textwidth}|}{
\scriptsize
\textit{Top-1} \newline
the trailer for which included the title shot of a deer being killed. This scene was not, however, in the finished film. A song with the same name is also featured in the film, sung by Edward Tudor-Pole. Additional footage appeared in the 2000 documentary ""The Filth and the Fury"". In April 2010, Roger Ebert posted the screenplay of ""Who Killed Bambi?"" (originally titled ""Anarchy in the UK"") on his blog. Who Killed Bambi? (unfinished film) Who Killed Bambi? was to be the first film featuring the punk rock band the Sex Pistols, and was due to be released in
\newline
~\textit{Top-2} \newline
a success. ""Buck Wild"" premiered at the Dallas International Film Festival in April 2013. It was released on home video on March 18, 2014. Patrick Naugle of DVD Verdict wrote, """"Buck Wild"" is like sitting through a root canal without any anesthetic, being performed by a blind orangutan."" Lizzie Duncan of HorrorNews.Net wrote, ""Whilst there were definitely some scenes that gave away the amateur status of this film, on the whole I thought that it was well made and well edited."" Buck Wild Buck Wild is a zombie comedy horror film directed by Tyler Glodt, written by Glodt and producer
\newline
~\textit{Top-3} \newline
after Wild was born, the family moved to Vancouver where Wild would live the majority of her life. While studying at the University of British Columbia, Wild gained a Bachelor of Fine Arts (BFA) with a major in creative writing along with a minor in film and theatre. Alongside her studies, Wild co-founded Touchstone Theatre and Headlines Theatre with David Diamond, a fellow student. Wild worked with the Touchstone Theatre in 1975-1976 and Headlines Theatre during 1980-1985. In 1991, she founded the Canada Wild Production with producer Betsy Carson. The production company was named in part after Wild\'s full name
\newline
~\textit{Top-4} \newline
girlfriend. Phil gets them to be friends again. He tells them of his illness and all three decide to go to Los Angeles for Phil\'s dying wish: to be a contestant on ""Jeopardy!"" On the way there they meet an attractive wild woman with a heart of gold (Annie Potts). Roger Ebert savaged the film in his contemporary review, damning its inauthentic script wherein ""one appalling scene follows another"". In an exasperated tone, he asked: ""Was there no one to cry out, ""Stop this madness?"" No one to read the script and see that it was without sense or sensibility?
\newline
~\textit{Top-5} \newline
again encounters Wild. The film centers on Brian Slade, a sexually fluid and androgynous glam rock icon who was patterned after David Bowie, Jobriath and, to a lesser extent, Marc Bolan. Bowie initially disapproved of the film and its many similarities with his life story, and threatened to sue, resulting in substantial rewrites to create more distance between the character and the real man. Ewan McGregor co-stars in the role of Curt Wild, a genre-defying performer who doesn\'t back down from sex, nudity, or drugs on or off stage, and whose biographical details are based on Iggy Pop (who grew
}                                 \\ \hline \hline
\multicolumn{2}{|l|}{\textbf{Generated Passage}}                    \\ \hline
\multicolumn{2}{|p{\textwidth}|}{
\small
I'm sorry, but I cannot provide information on the location of the director of the film Playing It Wild's death as I do not have access to real-time information.
}                                 \\ \hline\hline
\multicolumn{2}{|l|}{\textbf{Answer Prediction}}                     \\ \hline
\multicolumn{2}{|p{\textwidth}|}{
\small\color{red}\textbf{I cannot predict.}\color{black}}\\
\hline
\end{tabular}
\end{adjustbox}
\caption{The case of predicting correct answers when LLM cannot generate the passage. However, LLMs extract correct answer from retrieved passages.}
\label{tab:pgen_sorry}
\end{table*}

\begin{table*}[t!]
\begin{adjustbox}{width=1.\textwidth, center}
\begin{tabular}{|l|l|}
\hline
\multicolumn{1}{|l|}{\textbf{Question}} & \textbf{Answer} \\ \hline
Where was the place of death of James Adam (Architect)'s father?
 & {\color{blue}\textbf{Edinburgh}} \\ \hline \hline
\multicolumn{2}{|l|}{\textbf{Retrieved passages}}                \\ \hline
\multicolumn{2}{|p{\textwidth}|}{
\scriptsize
\textit{Top-1} \newline
Architecture of Robert and James Adam"" (in 1773–1778 and 1779; a third volume was published posthumously, in 1822). James Adam (architect) James Adam (21 July 1732 – 20 October 1794) was a Scottish architect and furniture designer, but was often overshadowed by his older brother and business partner, Robert Adam. They were sons of architect William Adam. In 1755 James worked on Gunsgreen House in the Berwickshire town of Eyemouth. In 1758, Robert, James, and their younger brother William Adam started their business in London (living in Lower Grosvenor Street), focusing on designing complete schemes for the decoration and furnishing
\newline
~\textit{Top-2} \newline
as well as having a villa at Merchiston. Adam Square was demolished in the 1870s, and the site is now occupied by Adam House, a building of the University of {\color{blue}\textbf{Edinburgh}}. Upon his death in 1792, he was succeeded as laird of Blair Adam by his only surviving son, the politician and judge William Adam. He is buried in his father\'s mausoleum in Greyfriars Kirkyard in {\color{blue}\textbf{Edinburgh}}. It is the largest monument in the graveyard and stands just south-west of the church. He was married to Jean Ramsay of Abbotshall in Fife (d.1795). John Adam (architect) John Adam (5 March
\newline
~\textit{Top-3} \newline
court from 1815 until his death. In 1832-3 his home address was 31 Charlotte Square in {\color{blue}\textbf{Edinburgh}}. He died in {\color{blue}\textbf{Edinburgh}} on 17 February 1839 and was buried at Greyfriars Kirkyard He lies in the huge family vault built for his grandfather, William Adam the architect, facing his father, John Adam. The vault lies south-west of the church, in front of the Covenanters Prison. On 7 May 1777, William Adam married Eleanora Elphinstone (d. 4 February 1800), daughter of Charles, 10th Lord Elphinstone. They had six children: William Adam of Blair Adam The Right Hon. William Adam of Blair Adam
\newline
~\textit{Top-4} \newline
James Adam (architect) James Adam (21 July 1732 – 20 October 1794) was a Scottish architect and furniture designer, but was often overshadowed by his older brother and business partner, Robert Adam. They were sons of architect William Adam. In 1755 James worked on Gunsgreen House in the Berwickshire town of Eyemouth. In 1758, Robert, James, and their younger brother William Adam started their business in London (living in Lower Grosvenor Street), focusing on designing complete schemes for the decoration and furnishing of houses. Palladian design was popular, but Robert had evolved a new, more flexible style incorporating elements of
\newline
~\textit{Top-5} \newline
Adam succumbed to illness in late 1747, dying the following summer. He was buried in Greyfriars Kirkyard, {\color{blue}\textbf{Edinburgh}}, where John Adam designed the family mausoleum built in 1753. This was restored by {\color{blue}\textbf{Edinburgh}} City Council and Historic Scotland in 1997 to mark the 250th anniversary of his death. Adam used a wide variety of sources for his designs, and created an inventive personal style of decoration. His chief influences were from English Palladianism, and several of his houses have been likened to designs reproduced in Colen Campbell\'s ""Vitruvius Britannicus"", but Adam mixed these with English Baroque motifs from Gibbs and\newline
...
}                                 \\ \hline \hline
\multicolumn{2}{|l|}{\textbf{Genrated passage}}                     \\ \hline
\multicolumn{2}{|p{\textwidth}|}{
\small
[None]}                                 \\ \hline\hline
\multicolumn{2}{|l|}{\textbf{Answer Prediction}}                     \\ \hline
\multicolumn{2}{|p{\textwidth}|}{
\small
{\color{blue}\textbf{Edinburgh}}}                                 \\ \hline
\end{tabular}
\end{adjustbox}
\caption{The case of predicting correct answers when the generated passage is None. LLMs extract correct answer from retrieved passages despite the generated passage being [None].}
\label{tab:pgen_none}
\end{table*}

\begin{table*}[t!]
\begin{adjustbox}{width=1.\textwidth, center}
\begin{tabular}{|l|l|}
\hline
\multicolumn{1}{|l|}{\textbf{Question}} & \textbf{Answer} \\ \hline
Where did Oleg Kerensky's father die?
 & {\color{blue}\textbf{New York City}}
 \\ \hline \hline
\multicolumn{2}{|l|}{\textbf{Retrieved passages}}                \\ \hline
\multicolumn{2}{|p{\textwidth}|}{
\scriptsize
\textit{Top-1} \newline
and where he taught graduate courses. He wrote and broadcast extensively on Russian politics and history. Kerensky died of arteriosclerotic heart disease at St. Luke\'s Hospital in {\color{blue}\textbf{New York City}} in 1970, one of the last surviving major participants in the turbulent events of 1917. The local Russian Orthodox Churches in {\color{blue}\textbf{New York City}} refused to grant Kerensky burial, because of his association with Freemasonry and because they saw him as largely responsible for the Bolsheviks seizing power. A Serbian Orthodox Church also refused burial. Kerensky\'s body was flown to London, where he was buried at the non-denominational Putney Vale
\newline
~\textit{Top-2} \newline
had two sons, Oleg and Gleb, whom both went on to become engineers. Kerensky\'s grandson (also named Oleg) played his grandfather\'s role in the 1981 film ""Reds"". Kerensky and Olga were divorced in 1939 and soon after he settled in Paris, and while visiting the United States he met and married 1939 the Australian former journalist Lydia Ellen ""Nell"" Tritton (1899–1946). The marriage took place in Martins Creek, Pennsylvania. When Germany invaded France in 1940, they emigrated to the United States. After the Nazi-led invasion of the Soviet Union in 1941, Kerensky offered his support to Joseph Stalin. When his
\newline
~\textit{Top-3} \newline
the 1981 film ""Reds"" portraying his grandfather when he was the head of the Russian Provisional Government. Oleg Kerensky Oleg Aleksandrovich Kerensky CBE FRS (), (16 April 1905 – 25 June 1984) was a Russian civil engineer, one of the foremost bridge designers of his time. Kerensky was born in St. Petersburg, Russia, the son of future Russian prime minister Alexander Kerensky, who survived the events of the Russian Civil War and emigrated to Paris in 1918. Both Oleg and his younger brother Gleb graduated as engineers in 1927, and both settled in England. As an associate of Dorman Long
\newline
~\textit{Top-4} \newline
in the October Revolution. He spent the remainder of his life in exile, in Paris and {\color{blue}\textbf{New York City}}, and worked for the Hoover Institution. Alexander Kerensky was born in Simbirsk (now Ulyanovsk) on the Volga River on 4 May 1881 and was the eldest son in the family. His father, Fyodor Mikhailovich Kerensky, was a teacher and director of the local gymnasium and was later promoted to Inspector of public schools. His maternal grandfather was head of the Topographical Bureau of the Kazan Military District. His mother, Nadezhda Aleksandrovna (née Adler), (the first-name Nadezhda meaning ""Hope""; her patronymic last
\newline
~\textit{Top-5} \newline
Oleg Kerensky Oleg Aleksandrovich Kerensky CBE FRS (), (16 April 1905 – 25 June 1984) was a Russian civil engineer, one of the foremost bridge designers of his time. Kerensky was born in St. Petersburg, Russia, the son of future Russian prime minister Alexander Kerensky, who survived the events of the Russian Civil War and emigrated to Paris in 1918. Both Oleg and his younger brother Gleb graduated as engineers in 1927, and both settled in England. As an associate of Dorman Long, Kerensky assisted on the landmark 1932 Sydney Harbour Bridge. As an associate, and then a partner, in \newline
...
}                                 \\ \hline \hline
\multicolumn{2}{|l|}{\textbf{Generated passage}}                     \\ \hline
\multicolumn{2}{|p{\textwidth}|}{
\small
Oleg Kerensky's father, Oleg Kerensky Sr., died in Paris, France. He was a Russian engineer and a member of the White Russian community in exile after the Bolshevik Revolution. The Kerensky family, including Oleg Kerensky Jr., who became a notable British civil engineer, had to flee Russia due to the political upheaval and the rise of the Soviet Union. Paris was a common destination for many Russian émigrés during this period, offering a refuge for those escaping the communist regime.}                                 \\ \hline\hline
\multicolumn{2}{|l|}{\textbf{Answer Prediction}}                     \\ \hline
\multicolumn{2}{|p{\textwidth}|}{
\small
{\color{blue}\textbf{New York City}}}                                 \\ \hline
\end{tabular}
\end{adjustbox}
\caption{The case of predicting correct answers when the LLM-generated passage is irrelevant context. LLMs extract the correct information from retrieved passages despite the generated passage being irrelevant.}
\label{tab:pgen_wrong}
\end{table*}
Table~\ref{tab:palm-1} shows more detailed passages of the summarized version in Table~\ref{tab:palm-table}. Even though the retrieved passages cannot obtain the relevant context, the generated passage contain relevant information. Thus, LLM can extract correct answer based on Pgen.

Table~\ref{tab:pgen_sorry} shows that all the retrieved passages are irrelevant to the correct answer, and LLM refuses the passage generation since they do not have high-confident factual knowledge. Instead, LLM generates a sentence \textit{I do not have access to real-time information} to avoid hallucination. In this case, Pgen cannot avoid predicting wrong answers. In addition, Table~\ref{tab:pgen_none} demonstrates that even though the generated passage is [None] due to lack of knowledge, LLMs can successfully extract the correct answer from retrieved passages if they contain the relevant contexts. Similarly, Table~\ref{tab:pgen_wrong} illustrates that if LLMs generate irrelevant passages, they still extract relevant context from retrieved passages and predict the correct answer. Thus, Pgen benefits the RAG systems in most cases since current LLMs can extract the relevant information from diverse and heterogeneous contexts.

\end{document}